# Handling temporality of clinical events with application to Adverse Drug Event detection in Electronic Health Records: A scoping review

Maria Bampa[1]

[1]Karolinska Institute, Dept. of Learning, Informatics, Management &Ethics, bampa.maria@stud.ki.se

**Abstract**

*The increased adoption of Electronic Health Records(EHRs) has brought changes to the way the patient care is carried out. The rich heterogeneous and temporal data space stored in EHRs can be leveraged by machine learning models to capture the underlying information and make clinically relevant predictions. This can be exploited to support public health activities such as pharmacovigilance and specifically mitigate the public health issue of Adverse Drug Events(ADEs). The aim of this article is, therefore, to investigate the various ways of handling temporal data for the purpose of detecting ADEs. Based on a review of the existing literature, 11 articles from the last 10 years were chosen to be studied. According to the literature retrieved the main methods were found to fall into 5 different approaches: based on temporal abstraction, graph-based, learning weights and data tables containing time series of different length. To that end, EHRs are a valuable source that has led current research to the automatic detection of ADEs. Yet there still exists a great deal of challenges that concerns the exploitation of the heterogeneous, data types with temporal information included in EHRs for predicting ADEs.*

**Keywords:**

Adverse Drug Events, Electronic Health Records, temporal abstraction, data mining

## Introduction

Adverse drug events (ADEs) are injuries that are related to drugs and are caused by medication errors, allergic reactions or overdoses. [1] Each year they account for 1 million emergency department visits and 125.000 hospital admission in the USA.
[1] Those frequently occur in the hospital setting, causing besides human suffering a burden in the health care sector. [1] Prolonged hospitalizations, increased hospital admissions and additional use of health care resources could be the cause of that burden. [1] Drug safety is evaluated on clinical trials before launching to market, but that does not imply that all possible implications are measured, resulting later in medication withdrawal. Therefrom, it is of foremost importance to reduce the impact of ADEs, as that will lead healthcare to a safer and more sustainable future.

A relevantly new way to mitigate the problem of ADEs is by exploiting Electronic Health Record (EHR) data. EHRs are a valuable source of information as they do not only provide a holistic overview of patient's medical history, but they are also the source of longitudinal data, recorded chronologically. [2] Exploiting temporal data that are not evenly distributed over time is a promising alternative for detecting ADEs, yet challenging for standard machine learning algorithms.[3] When ignoring the temporal side of data, each case is described by values most recent in time, therefore leading to a great loss of information.

On the contrary, considering the fact that for each patient case, a specific feature consists of multiple recording thus, including temporality as a parameter, opens up great opportunities for the effective prediction of ADEs.

Temporal modelling on EHR data can be a challenging task when taking into account multivariate asynchronously sampled features, which makes classic time series classification techniques and features extraction methods not easily applicable.
[4] It is apparent, that temporal information needs to be handled differently from static attributes. There has been active work on learning classification models from multivariate time series data.

The aim of this article is, therefore, to investigate various ways of handling temporality in EHR data for ADE identification and detection. Based on selected literature, an overview and comparison of the most current methods is provided, as well as some directions for future research.

The above lead to the following research questions:

- How can the temporal aspect of the feature space that stems from EHRs be exploited for Adverse Drug Event detection?
    o How can the predictive performance of the chosen classifiers get improved by temporal pattern mining in complex multi-variate series data?
    o What are the time windows that should be considered when exploiting the temporal features of EHRs for ADE?

## Methods

1. Sources

Data mining in EHRs is an interdisciplinary field of study, covering studies from both data science and medical background. That is why this review is performed on databases that cover both these fields. The aforementioned databases are: Pubmed, Web of Science, as well as Google Scholar. In order to obtain the necessary results a combination of the following keywords was used: [temporal abstraction] AND [Electronic Health Records] AND [Adverse Drug Events], [machine learning] AND [prediction] AND [electronic health records], [data mining] AND [adverse drug events].

Articles published in any other language than English were excluded from the study. Also, the study does not include no peer reviewed articles. Last but not least, articles published before the year 2008 were also excluded considering that the field covered in this review is rapidly changing over the years.

To study the most recent development in the field, original articles, published the last 10 years, were included in the study. To continue with, in order to get an overview of the development in the field review articles regarding temporal mining in Electronic Health Records were also included, but studied separately. Studies regarding temporal mining, temporal abstractions and mining of temporal features for classification in EHRs were included regardless of focusing on detection of ADEs or not. That being so, the scope of the study was limited to the detection of adverse drug events from EHRs but went beyond the scope of ADEs in order to discover the advances in a field closely related prediction of diseases or conditions such as ADEs.

2. Study selection

For the first selection, the abstracts of all the articles obtained were studied to decide whether they are within the scope of the study. Articles irrelevant to Electronic Health Records or prediction of ADEs or classification of temporal features and classification of multivariate series were excluded.

The decision upon relevance was made according to the following criteria:

- Firstly, the focus was on including articles relevant to the classification of multivariate time series and the abstraction of temporal features in Electronic Health Records.
- Secondly, there should be a clear connection of handling temporal features from EHRs with the purpose of predicting a patient's disease or condition.

After the initial selection, a total of 11 articles were chosen as highly relevant to this review.

Amount of hits per database

| Search Words | Google Scholar | Web of Science | PubMed |
|---|---|---|---|
| EHR AND temporal abstraction | 40400 | 19 | 15 |
| ADEs AND EHRs AND temporal mining | 44800 | 23 | 20 |
| Time series classification AND ADEs AND EHR | 99000 | 7 | 13 |

## Results

In recent years, temporal mining has gained increased attention in the field of data mining and health informatics. A step towards a safer healthcare is to understand how events can evolve in time and how these can be exploited in the correct way to increase the predictive performance of classification methods.

The first part of the results section presents an overview of various ways generating and handling temporal features that represent time series. According to the literature retrieved the main methods were found to fall into 5 different approaches: Those that are based in temporal abstraction, those that transform data points into time series intervals, based in graph phenotyping, those that learn temporal weights and last but not least those that utilize data tables containing time series of different length.

1. Build on temporal abstraction

Batal et al. in their paper 'A Temporal Pattern Mining Approach for Classifying Electronic Health Record Data' rely on temporal abstractions and temporal pattern mining for the extraction of the features that belong to EHR data. [4] The purpose of their paper is to tackle the problem of the large number of temporal patterns being returned from the temporal mining task. They propose a language to describe temporal interactions among multiple time series. [4] They adopt the Minimal Predictive Temporal Patterns (MPTP) framework to predict patients who are at risk of developing heparin-induced thrombocytopenia. Their objective is to learn a function that can predict risk patients, therefore they apply a transformation that maps each EHR instance to a fixed size feature vectors while preserving their temporality. [4] They also experiment with changing the window size from 5 fixed days to varying from 3 to 9 days of the patient record. They concluded that the most recent temporal patterns have higher predictive performance. [4]

A slightly different direction is proposed in the paper 'Mining Recent Temporal Patterns for Event Detection in Multivariate Time Series Data' by Batal et al., where they take into account the temporality of the feature space and they mine the temporal patterns backwards in time by applying the Recent Temporal Pattern mining framework(RTP). [5]

Advantages of their framework are some of the following: RTP focuses on temporal patterns that are more useful for classification, frequent RTPs is much smaller than the number of frequent temporal patterns and last but not least the mining algorithm is very efficient especially when it comes to large databases. [5]

On that study, they suppose that clinical measurements that are close to the target event are more predictive than others more distant in time. As opposed to the first study [4] where they present an algorithm that utilizes the most frequent temporal patterns separately for each label.

Patel et al. in their paper 'Mining Relationships Among Interval-based Event Classification' are expanding Allens' interval algebra [6] from representing just two interval-based events to capturing temporal relationships among three or more interval-based events. [7] They propose an interval-based mining algorithm to reduce the search space and eliminate not necessary candidates. To make the presentation lossless, they used additional information to augment the hierarchical representation. [7] After applying their algorithm to a real dataset they concluded that the predictive performance of closely related classes is improved. [7]

2. Time-stamped data points into symbolic time series intervals

A different approach is, to transform data points into time intervals to mine the frequent time interval related patterns that will be used for prediction. The following papers are using this approach.

To start with, Moskovitch and Sahar in their paper 'Classification of multivariate time series via temporal abstraction and time intervals mining' propose the transformation of the time point time series into time series interval representation, to analyze the relationship between variables. [8] Their approach does not make any assumptions regarding the data. [8] They move towards the transitive property of temporal relations and they introduce KarmaLego for candidate generation. [9]

The next paper by Moskovitch et al. is trying to introduce an improvement regarding the KarmaLego algorithm. KarmaLego algorithm can handle thousands of symbols, but every study until then, including the study mentioned above, have not utilized that, instead they used only a small number of symbols. [9] That being the case they introduced an improvement of KarmaLego for the discovery of frequent patterns that can handle thousands of symbols. [9] They based their model on EHR data and diagnoses and they concluded that using advanced TIRP representation is better than using TIRPs with boolean representation. [9]

3. Graph-based electronic phenotyping

A rather different direction is to handle the temporality of data using temporal graphs. The proposed novel framework in this section is representing a patient's EHR as a graph, where the nodes represent the medical events and the edges encode temporality amongst medical those. [10] That temporality will be measured with assigned weights on the edges that represent duration between two-time events. They validate their method with data of patients with risk of heart failure and patients with COPD pre-condition. They mention that, they address the problem of irrelevant pattern explosion and that all the detected phenotypes are represented as graphs instead of subsequences thus, addressing also the problem of interoperability of the mined phenotypes. [10]

4. Learning temporal weights

There have been some attempts to handle the temporality of data in EHRs using temporal weights. Some of them are presented below:

Zhao et al. propose three different strategies to handle the temporal aspect of the data for detecting ADEs: bag of events, bag of binned events and bag of weighted events. The best observed strategy was found to be the bag of weighted events, that assigns weights to events from different time windows and according to their distance from the target ADE. [11]

Another work by Zhao investigates 9 different temporal weighting strategies that assign weights according to a curve function f(n), for an event that happened n days before the appearance of an ADE. [12] In that case the weights are assigned in a more universal and fixed way as compared to the first study.

Another contribution to that field comes from Zhao and Henriksson, where they extend the previous research [12] by learning weights of clinical events consider important the informative importance of the target event. [13] They suppose that higher weights should be given to events that are more informative and not only consider how recent they are to a specific clinical event.

5. Data tables containing time series of different length

Zhao et al. in their study 'Learning from heterogeneous temporal data in electronic health records' attempt to handle data series of different length that are stored in tabular format with application to ADEs from EHRs. [3] Each time series is transformed to the SAX representation that results in single valued features, in order to be easier handled by the classifier. They investigated techniques of transforming temporal and sequential information into table format for time series that differ in length.

## Discussion

The complex structure of EHR data poses a problem in the use of standard machine learning algorithms not only for detecting ADEs but for every kind of predictive task. Many approaches handle the data as static features which leads to great loss of information. Taking into account the temporality and heterogeneity of the feature space that stems from EHRs will lead to better predictive models for the task of identifying ADEs.

Several techniques have been proposed in the academic field for exploitation of sequential data, the most important of which are presented in this study. Different methods, from temporal abstractions and assigning weights to graph based techniques have been proposed, to tackle this issue. It is important to mention that these are different approaches using different kinds of datasets which makes the comparison a difficult task, so the decision lies to which one assists the classification methods to perform better.

Both papers from Batal et al. [4][5] are using temporal abstraction for the representation of temporal data. The former converts the time series into a higher lever description whereas the latter coverts them into time interval sequences. That is because on the one hand they take the frequency into

Batal et al. they even conclude that changing the window size to include more days increased the number of patterns obtained, which led to an increase of the execution time. [4]

To continue with, another interest is to handle large numbers of data by exploiting time series intervals and then discover frequent event patterns for knowledge discovery. Converting the data from time-stamped to a series of uniform time intervals can benefit the pattern mining process. [8] Both studies by Moskovitch et al. [8][9] are utilizing the symbolic time intervals for mining frequent patterns but the second paper is making an improvement regarding the number of symbols that KarmaLego can handle. In that case, the second enhanced algorithm option would perform better, because EHRs consist of multi-dimensional fields, for example diagnoses, procedures, drug codes etc. However, they only test their data with the purpose of predicting clinical procedures as outcome events using a time window of two years for each patient. It would be interesting to see, though, that method with application to ADE detection and with different time windows.

On the other hand, using graph-based temporal phenotyping as in the work of Liu et al. [10] could offer a more simplified version of handling temporality in EHRs. That version presents each patient as a graph where the nodes are the medical events and the edges represent temporality. The average duration between those events will be marked with a weight. That technique is applicable for the detection of temporal relationships between two events and not for the prediction of those. Also, among the large number of clinical events only a small number could be identified as informative for its relevance to an ADE. That being the case it could be less useful for prediction of ADEs.

The aspect of considering which time periods should be taken into account had been studied in the papers [11], [12], [13]. An ADE could be caused by a combination of reasons in different time points. Therefore, it is more efficient regarding detection of adverse drug events not only to take temporality into account but also to check how informative is the event according to the target, as they did in study [13].

Last but not least, the study by Zhao and Henriksson, for detecting ADEs using data series of different length stored in table format suffers from this limitation: the resulted subsequence does not need to adhere to the description of the original sequence. Therefore, resulting in subsequences that can be more or less informative than the previous ones.

- Limitations of this study

This study is a scoping review so the research that was conducted here was not thorough and some important contributions might have been missed. That can also be attributed to the fact that only English literature was included in the study and to time constraints for the preparation of the survey.

**Conclusions**

While there is no clear answer on which method of handling temporal features is better, it is apparent that sequential data can lead to a better understanding of ADEs and improve their detection. One could conclude that the best answer lies in the nature of the dataset itself, how it is represented and what is the predictive task. According to the literature reviewed, taking into account the temporality of data and not just handling them as static features increases the predictive performance of each classifier used. The question of what time windows should be considered for the prediction of ADEs is rather complicated. ADEs are highly dependent on the point of time which the medication was taken. Therefore, a potential decrease of the time window might result in great loss of information while an increase of the time window can cause a retrieval of a great amount of not so informative data.

Conclusively, it is of great importance to direct future research on exploiting temporal features for the detection of adverse drug events, as most of the studies retrieved concerned matters of predicting a wide variety of other medical conditions. A focus could be on utilizing both static and temporal features separately and in conjunction and investigate their importance towards ADE identification and prevention. It would also be very interesting to investigate the effect in predictive performance of the chosen classifiers, of a possible change of the patient window size length.